\pdfoutput=1

\documentclass[11pt]{article}

\usepackage[final]{coling}
\usepackage{xcolor}
\usepackage{ulem}
\usepackage[inkscapelatex=false]{svg}
\usepackage{times}
\usepackage{adjustbox}
\usepackage{multirow}
\usepackage{latexsym}

\usepackage[T1]{fontenc}

\usepackage[utf8]{inputenc}

\usepackage{microtype}

\usepackage{inconsolata}

\usepackage{graphicx}

\title{LLM-Friendly Knowledge Representation for Customer Support}

\author{Hanchen Su \quad Wei Luo \quad Wei Han \quad Yu Liu \\
        \quad \textbf{Yufeng Zhang} \quad \textbf{Cen Zhao} \quad \textbf{Joy Zhang} \quad \textbf{Yashar Mehdad}\\
  Airbnb Inc., USA \\
  \texttt{\{hanchen.su, wei.luo, wei.han, elaine.liu} \\
  \texttt{wayne.zhang, mia.zhao, joy.zhang, yashar.mehdad\}@airbnb.com}
}

\begin{document}
\maketitle
\begin{abstract}
We propose a practical approach by integrating Large Language Models (LLMs) with a framework designed to navigate the complexities of Airbnb customer support operations. In this paper, our methodology employs a novel reformatting technique, the Intent, Context, and Action (ICA) format, which transforms policies and workflows into a structure more comprehensible to LLMs. Additionally, we develop a synthetic data generation strategy to create training data with minimal human intervention, enabling cost-effective fine-tuning of our model. Our internal experiments (not applied to Airbnb products) demonstrate that our approach of restructuring workflows and fine-tuning LLMs with synthetic data significantly enhances their performance, setting a new benchmark for their application in customer support. Our solution is not only cost-effective but also improves customer support, as evidenced by both accuracy and manual processing time evaluation metrics.
\end{abstract}

\section{Introduction}
Customer support at Airbnb aims to assist users in resolving a wide range of issues throughout their journey. The effectiveness of service delivery relies on a thorough understanding of Airbnb-specific knowledge, including policies, workflows, and troubleshooting manuals. Airbnb agents leverage specialized training and cognitive skills to apply this knowledge to resolve customers issues. As a result, customer support is a complex challenge in Airbnb.

\begin{figure}
    \centering
    \includegraphics[width=1\linewidth]{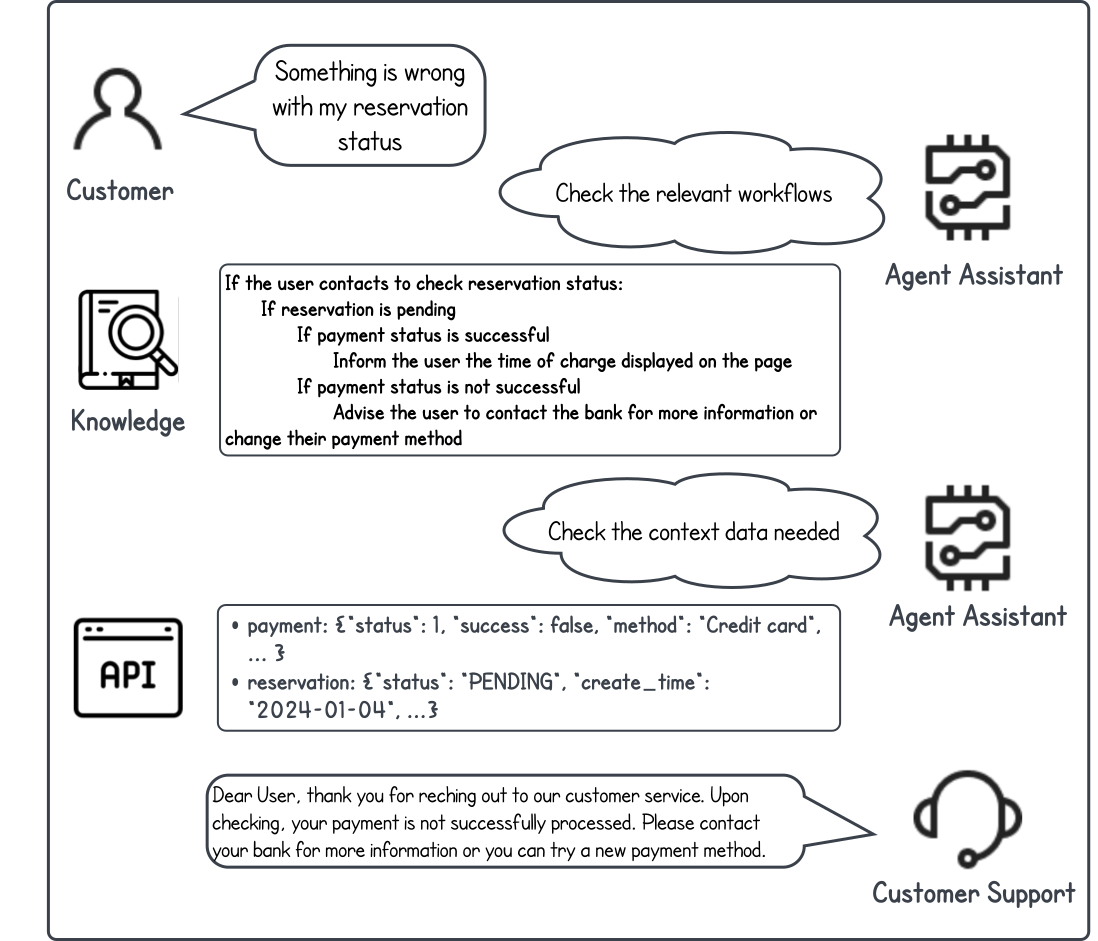}
    \caption{Intelligent customer support: generate the correct response based on internal workflows and context data}
    \label{fig:1}
\end{figure}

In recent years, the fast development of Large Language Models (LLMs) provides technical breakthroughs that can scale and automate workflows in solving complex problems. It not only enhances automation efficiency but also allows human agents to focus on more complex and sensitive issues, optimizing the allocation of resources and improving agent's productivity and overall customer satisfaction.

Figure \ref{fig:1} illustrates a typical application in assisting a customer support application: when a customer poses a question, the system automatically retrieves relevant context information and workflows tailored to the user query and intent. The LLM within the system then uses this context, along with other available information, to generate appropriate responses. Although this approach may seem straightforward on the surface, developing an effective solution entails significant complexities to address.

\noindent \textbf{The Complexity of Internal Policy and Workflow Documents} These documents are typically written in complex terminology and technical jargon that requires special training for human agents to understand and follow. They often consist of lengthy, colloquial text with convoluted workflows that may not be mutually exclusive, which results in difficulty for LLMs to parse, understand, and reason over.
To ensure that an LLM can effectively interpret business knowledge, they need to be reformulated into a format that is digestible by LLMs (i.e., LLM-friendly). This rewriting and editing process requires the expertise and domain knowledge of experienced human agents, leading to significant costs that are often prohibitively expensive.

\noindent \textbf{Limitations of Larger LLMs} Larger and higher-quality language models can be slower and more costly. This is important in enterprise application since latency and cost are two important factors in system design. Additionally, they do not have internal domain knowledge specific to enterprises' customer support and products.

\noindent \textbf{Training Data Creation} The process of collecting data for model training is complex and costly. In particular, \emph{implicit knowledge} which is critical to effective problem solving does not exist in explicit format. Primarily to reduce the operation cost, agents often do not fully document the knowledge used and contextual data checked during the resolution process.

To solve the above-mentioned problems, we propose an end-to-end solution for LLM-based workflow-driven customer support automation. The rest of the paper will focus on the two key areas of this solution:

\begin{itemize} 
    \item \textbf{ICA: LLM-friendly knowledge representation} To enhance the \emph{interpretability} and \emph{reasoning} accuracy of LLMs in customer support tasks, we propose a new format called Intent, Context, and Action (ICA) to simplify, structure and represent the business knowledge. 
    \item \textbf{Fine-tuning LLM to improve comprehension and reasoning over ICA} Following the effective trend of levering data augmentation approaches \cite{liu2024bestpracticeslessonslearned} and the power of Chain of Thought (CoT) \cite{wang2023selfconsistencyimproveschainthought}, we develop a synthetic data generation approach to create training data 
    with minimal human involvement. Subsequently, we utilize 
    this synthetic dataset to fine-tune our model, thereby enhancing our LLM's performance using in-domain knowledge.
\end{itemize}

Our internal experiments demonstrate that this combined strategy enhances the performance of LLMs in the customer support reasoning tasks. This solution is intended solely for exploratory purposes which is not, and will not be, applied to Airbnb products. However, we hope that our solution can help with developing AI Agents for other business domains tackling similar problems. 

\section{Related Work}

While knowledge simplification and content reformatting is a straightforward strategy to enhance the quality and interpretability of traditional ML models, there hasn't been a lot of work in simplifying knowledge and content reformatting for LLMs. Various types of text rewriting have been explored, including paraphrasing \cite{siddique2020unsupervised,xu2012paraphrasing}, style transfer \cite{riley2020textsettr,zhang2020parallel,reif2021recipe}, and sentence fusion \cite{mallinson2022edit5}. RewriteLM, an instruction-tuned large language model designed for cross-sentence text rewriting, was introduced by \cite{shu2023rewritelminstructiontunedlargelanguage}. \cite{zhang2024comprehensivestudyknowledgeediting} highlighted how knowledge editing can be utilized to implement factual updates with minimal impact on the model’s performance and flexibility across different knowledge domains. To our knowledge, our study is among the first to explore the transformation of unstructured, complex text workflows into pseudocode to enhance LLM performance in specific domain tasks.



To minimize the effort of human annotation, contemporary studies in synthetic data generation are focusing on leveraging LLMs for data augmentation. This includes the generation of instructions, input, and output examples directly from a language model, followed by the removal of any invalid samples prior to their utilization in fine-tuning the base model \cite{wang2022self}. Other notable contributions in this area include the work of \cite{he2019revisiting,xie2020self,huang2022large} who demonstrated the efficacy of incorporating synthetically generated data into training. \cite{schick2022peer} introduced the PEER methodology, which employs LLMs to infill missing data points that are subsequently used to train other models. 
The closest work to our synthetic data generation solution is STaR \cite{zelikman2022starbootstrappingreasoningreasoning} which leverages CoT to generate synthetic rationales and filters out those leading to wrong answers for fine-tuning LLMs to improve their reasoning. 

In the domain of customer support, the integration of generative AI, particularly through LLMs, promises significant improvements in efficiency and service quality\cite{wei2023COT}.
\cite{reinhardgenerative} identifies several customer support activities such as transferring, escalating, generations and retention, that can be enhanced by LLMs. In a practical application, \cite{brynjolfsson2023generativeaiwork} observed a significant productivity increase among a large number of customer support agents after introducing an LLM-based conversational assistant, specifically for novice and low-skilled employees. 
This further highlights a significant gap in the literature, underscoring the need for more empirical studies to demonstrate the practicality of LLMs in automating tasks to improve customer support productivity and to define the necessary requirements for the effective deployment of more advanced technologies.

\begin{figure}[t!]
\centerline{\includegraphics[width=1\linewidth]{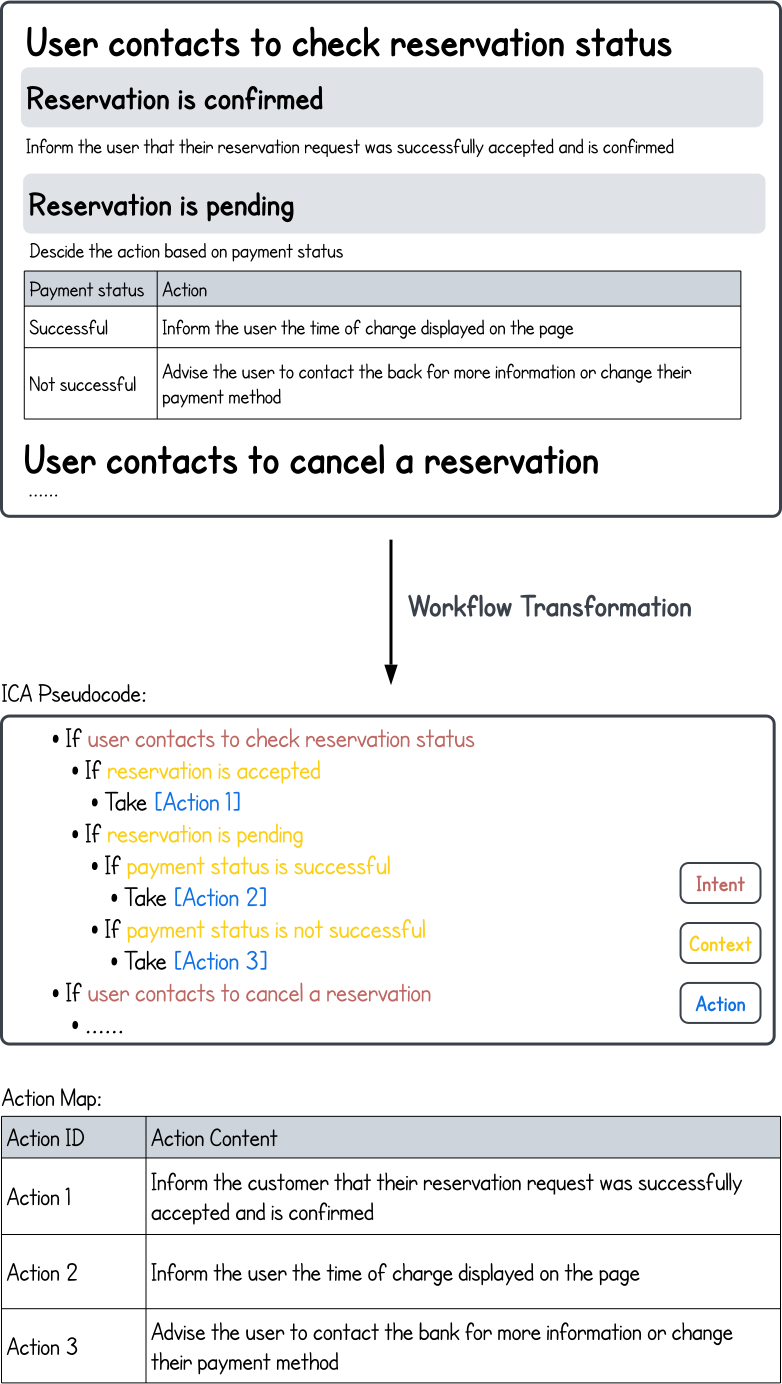}}
\caption{Converting workflows in one document from original (rich text) format to the ICA format.}
\label{fig: 2}
\end{figure}

\section{Business Knowledge Representation as ICA Pseudocode} \label{proposal}
Customer support problems can be treated as a Knowledge Base Question Answering problem (e.g., \citet{baek2023knowledge}): Given a user query and a knowledge base of workflows and policies, what is the correct response? This highlights the significance of business knowledge, along with the importance of their formatting and structure, in improving comprehension for both human agents and LLMs.

To this aim, we conducted an in-depth analysis of existing customer support workflows of Airbnb and identified a pattern of ``Intent, Contexts, Actions'' (ICA) which covers nearly all workflows that human agents needs to follow in order to respond to user queries. A typical workflow instructs when a user reaches out to customer support agent with a certain ``\textit{Intent}" (I), based on the conditions and the ``\textit{Contexts}" (C) of the user issue, what   ``\textit{Actions}" (A) human agents should take. These ICA workflow business knowledge are often defined by the business functions in explicit format, or ``implicit'' tribal knowledge based on experiences of human agents from solving similar issues in the past. 

Existing ``knowledge'' such as workflows are presented in a way that human agents can read and interpret with specific training and experience. These workflows are not consistently structured and not designed for LLMs to understand and interpret. Some workflows are represented as a mixture of structured information (e.g., hierarchy tree in instructions and tables) and unstructured data (e.g., text and image), stored in a rich text format including headings, markups, hyperlinks, lists, items, and tabular data with a complex textual descriptions of the conditions and policies around different actions and solutions. Tables, for example, are compact representations for trained human agents to find information (in a cell) associated to its corresponding row/column headers which are more challenging for LLMs to digest and interpret.


\begin{figure*}[t!]
\centerline{\includegraphics[width=1\linewidth]{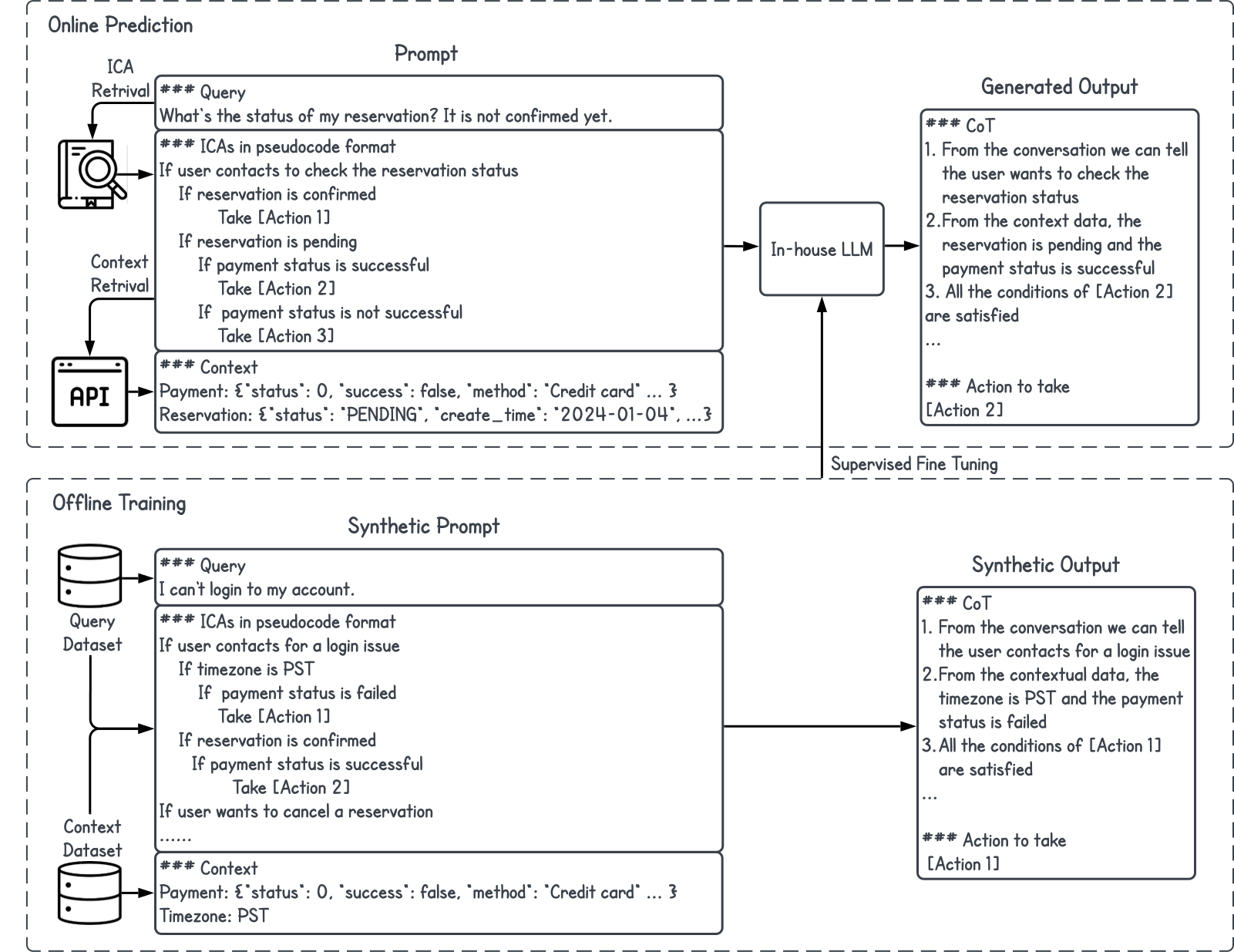}}
\caption{Our solution includes: 1) Transforming the workflow into ICA format, thereby enhancing the interpretive abilities of language models. 2) Online prediction: Retrieving relevant ICA candidates by comparing the user query and "Intent" part of the ICAs in the knowledge base; Retrieving necessary contextual data from backbend APIs; Utilizing LLMs to generate the action to take 3) Offline training: Addressing the scarcity of training data by employing synthetic methods to create the necessary data. We then apply Supervised Fine-Tuning (SFT) to train the open-source language models.}
\label{fig: 3}
\end{figure*}

In our approach, we propose to transform these workflows to the ICA structure as a pseudocode format. Our hypothesis is that by transforming the existing business knowledge to ICA format, we achieve a more LLM-friendly content, which is easier for LLMs to understand the business logic and to decide the right actions with higher accuracy. Figure \ref{fig: 2} shows a simplified example of the the original workflows in a rich text format and transformed as ICA.

Compared to traditional formal representations of business logic such as  programming language (e.g., Java, Python) or formal schema (e.g., json-style workflow), ICA style pseudocode is much easier to create (even by non-engineers) and  maintain. We will show in the rest of this paper that with the power of fine-tuned LLMs, AI Agents can now interpret and execute business logic defined in ICA format with higher accuracy. Note that details of human efforts and cost of translation and maintenance of ICA is described in Appendix \ref{app: ica transformation}.


Therefore, we reframe the problem addressed by LLMs as follows: given the business knowledge characterized as a set of ICAs, for a user query, infer the \textbf{I}ntent, select the appropriate \textbf{A}ction where \textbf{C}ontextual conditions are met, and generate corresponding responses. The remainder of the paper outlines our methodology for addressing this problem.

\section{Methodology}\label{section: Methodology}

To instruct LLMs to understand and interpret the ICA format, we need to 1) transform our customer support business knowledge to the ICA format, and 2) create a dataset to train (fine-tune) LLMs to learn how to interpret and reason over ICA. Figure \ref{fig: 3} shows the relationship of online prediction and synthetic data generation for LLM offline training. 

\subsection{Transforming Business Knowledge into ICA} \label{ICA}

We first process the existing workflows to decompose, extract and detect the type of the text contents from rich text format. Then, the extracted and detected Intent, Context and Action are represented in an intermediate decision tree that is further converted to pseudocode which can be reviewed and edited by knowledge writers of content and operation team. See Appendix \ref{app: tree} and \ref{app: ica transformation} for more details.

When transforming each workflow to ICA, we substitute the content of each action with an ID starting from 1, while preserving the mapping between the IDs to contents in an action map. Then in the training data synthesis, only the action ID is synthesized. In online action prediction, the LLM only generates the action ID and the actual content of the action is further queried from the action map. This approach provides multiple advantages: It improves the accuracy of the output by simplifying generated content and enabling a direct comparison between the action IDs generated by the LLM and the ground truth labels in evaluation, facilitating the acquisition of quantitative metrics for model training iteration. Additionally, it allows for a reduction in the token size of the prompt and the output for online prediction, thereby decreasing latency.

\subsection{Fine-tuning LLMs to Interpret ICAs through Synthetic Data} \label{synthetic data} 


We use a randomized synthesis method to generate supervised fine-tuning data format for our training data. One training instance consists of the user query, context data and candidate ICA workflows in pseudocode in the prompt, the CoT rationale \cite{wei2023COT} and action to take in the response. Our assumption is that the LLM can learn to understand the ICA format after being exposed to a vast amount of randomly generated data. Even though the synthesized data does not reflect the real business knowledge, the synthetic data is still effective in `teaching' the LLMs about the format. Figure \ref{fig: 4} shows the three-step process of synthesizing the training instance. See Appendix \ref{app: training data} for more details.

\begin{figure*}[t!]
\centerline{\includegraphics[width=1\linewidth]{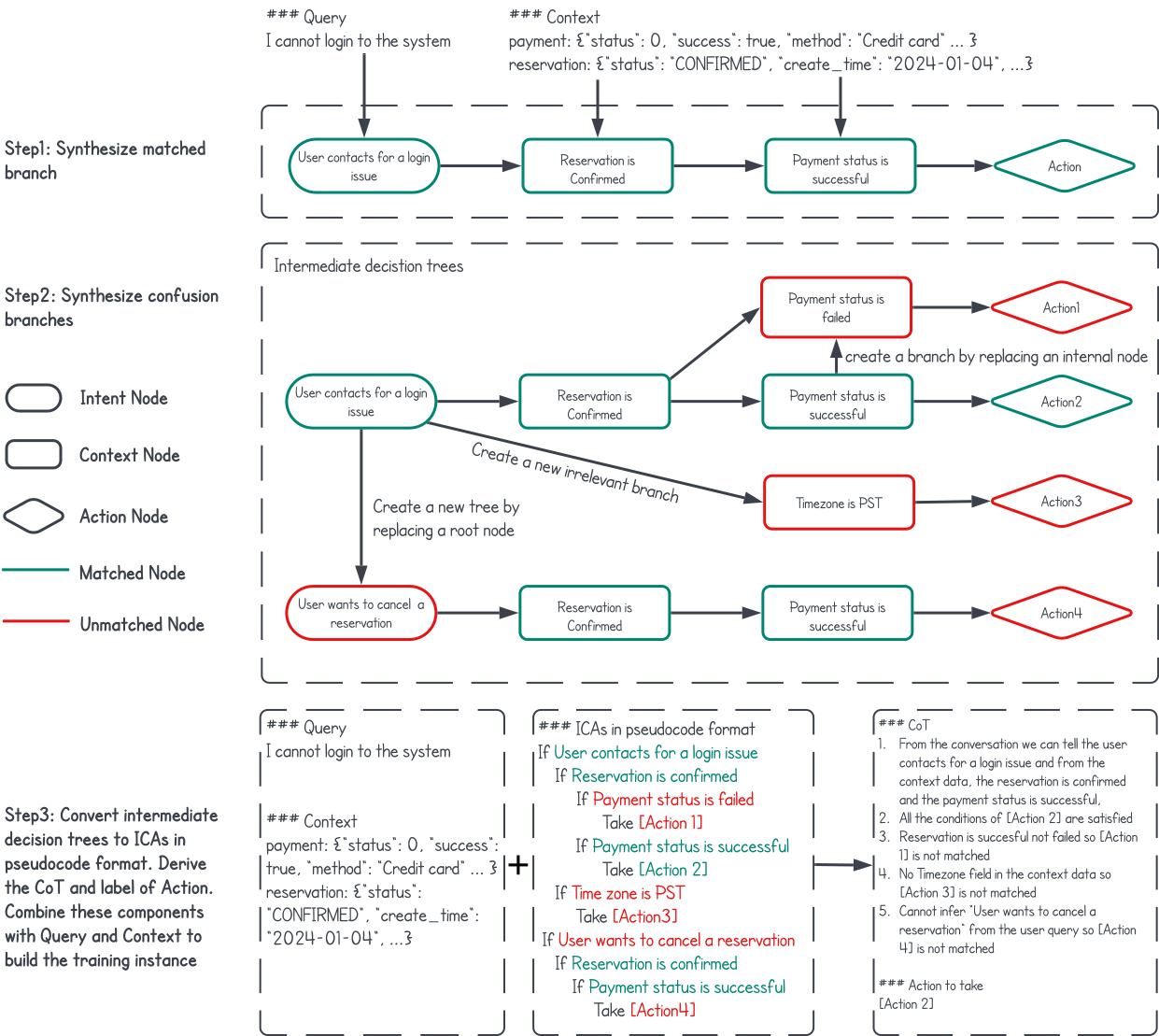}}
\caption{Three steps of generating synthetic training data: 1) Sample user query and context data randomly to establish a matched branch. 2) Incorporating additional divergent branches to construct the decision trees. 3) Developing pseudocode, detailing the reasoning process, and deriving the label from the trees, then integrating these components to assemble the training dataset.}
\label{fig: 4}
\end{figure*}

\section{Experiments}
\subsection{Experimental Setup}
We conducted a series of internal offline and online experiments to evaluate the quality and effectiveness of our proposed approach. In the online experiments, we predict actions to take derived from our methodology, as outlined in the paper, as a recommendation for agents to solve customer support inquiries based on the current conversation between the customer and the agent. Agents and cases were randomly assigned to control and treatment groups, with each group managing over 5,000 case assignments to ensure sufficient statistical power. To maintain focus on our primary research objectives and control for extraneous variables, we standardized the knowledge retrieval process across all experimental groups. This involved mapping user intentions to the top three most relevant original knowledge articles, which were then translated into ICA formats appropriate for each group. For each experimental group, models are selected based on their performance measured by offline evaluation metrics calculated from a dataset comprised of customer support conversations between customers and agents. Each instance in the dataset is labeled by human annotators with the appropriate action to take. 



\noindent \textbf{Evaluation Metrics} For offline evaluation, we use Accuracy (ACC) calculated as the number of cases with correct action prediction divided by the total number of the evaluation dataset. We also measure Average Latency (AL) based on the average time required to produce a response for each data point. 
For online evaluation, we use Average Manual Processing Time (AMPT) to evaluate the productivity of our solution. AMPT indicates the time spent to solve a case manually in different experimental settings. This metric indicates the effectiveness of our approach in saving time through reduced manual efforts which is directly linked to operational cost.
ACC and AL directly affect the overall performance: inaccurate suggestions can mislead agents, resulting in erroneous solutions and prolonged customer interactions. Latency will affect the agent waiting time for the suggestions and longer waiting time can result in a negative impact in customer satisfaction and also agent efficiency.

\noindent \textbf{LLMs} We selected two anonymized larger LLMs:  Model 1 and Model 2, along with smaller LLMs: Mixtral-8x7B-Instruct-v0.1 (Mixtral-8x7B) and Mistral-7B-Instruct-v0.2 (Mistral-7B) for comparison. Model training and serving details are documented in Appendix \ref{app: model training}.
 

\subsection{Experimental Results}
\begin{table}[ht!]
\centering
\begin{adjustbox}{max width=\textwidth}
\begin{tabular}{c|c|ll}
\hline
\multirow{2}{*}{\textbf{Model}} & \multirow{2}{*}{\textbf{CoT}} & \multicolumn{2}{c}{\textbf{ACC}}\\ \cline{3-4}
&& \textbf{Rich Text} & \textbf{ICA}\\
\hline
\hline
\multirow{2}{*}{Model 1} & w/o & 0.57 & 0.70 (\textbf{+0.13})\\
& w/ & 0.65 (\textbf{+0.08}) & 0.92 (\textbf{+0.25})\\
\hline
\multirow{2}{*}{Model 2}& w/o & 0.55 & 0.67 \textbf{(+0.12)}\\ 
& w/ & 0.61 (\textbf{+0.06}) & 0.89 (\textbf{+0.34})\\
\hline
Mixtral& w/o & 0.39 & 0.61 (\textbf{+0.22})\\ 
-8x7B& w/ & 0.43 (\textbf{+0.04}) & 0.82 (\textbf{+0.43})\\
\hline
Mistral& w/o & 0.16 & 0.51 (\textbf{+0.35})\\ 
-7B & w/ & 0.23 (\textbf{+0.07}) & 0.70 (\textbf{+0.54})\\
  \hline
  \hline
\end{tabular}
\end{adjustbox}
\caption{The ICA format and CoT can greatly enhance the model's accuracy with no fine-tuning.}
\label{tbl:1}
\end{table}
\noindent \textbf{ICA format and CoT can enhance the accuracy.} Table \ref{tbl:1} illustrates the impact of ICA format and CoT on the accuracy of models. Taking Model 1 as an instance, the baseline accuracy without the application of CoT format or ICA stands at 57\%. By employing ICA format alone, we observe a 13\% enhancement in accuracy. An additional 8\% increase is achieved through the utilization of CoT, and a substantial improvement of 25\% is realized when both ICA format and CoT are applied concurrently. The result also indicates that the ICA format consistently improves the accuracy of all models compared to the rich text format. This proves the effectiveness of our proposed ICA format for the knowledge representation in customer support applications. In addition, the results show that incorporating CoT further enhances the accuracy for all models in both rich text and ICA formats. This improvement is significantly more pronounced in the ICA format. Furthermore, we observe that the smaller models also exhibit notable accuracy gains with the ICA format and CoT, albeit starting from lower accuracy compared to larger LLMs. Among all LLMs, Model 1 demonstrates the highest quality when no fine-tuning is performed. 

\begin{table}[ht!]
\centering
\begin{adjustbox}{max width=\textwidth}
\begin{tabular}{c|c|c|c|c}
\hline
\multirow{2}{*}{\textbf{Model}} & \textbf{Fine-} & \multirow{2}{*}{\textbf{CoT}} & \multirow{2}{*}{\textbf{ACC}} & \multirow{2}{*}{\textbf{AL}} \\
&\textbf{Tuning}&&& \\
\hline
\hline
\multirow{2}{*}{Model 1} & \multirow{2}{*}{-} & w/o & 0.70 & 16.6s\\ 
&& w/ & 0.92 & 46.4s \\
\hline
\multirow{2}{*}{Model 2}& \multirow{2}{*}{-} & w/o & 0.67 & 15.9s\\ 
&& w/ & 0.89 & 44.2s \\
\hline
\multirow{4}{*}{Mixtral-8x7B}& \multirow{2}{*}{w/o} & w/o & 0.61 & 11.3s\\ 
&& w/ & 0.82 & 20.0s \\
&\multirow{2}{*}{w/}& w/o & 0.67 & 4.7s \\
&& w/ & 0.86 & 8.0s \\
\hline
\multirow{4}{*}{Mistral-7B}& \multirow{2}{*}{w/o} & w/o & 0.51 & 5.7s\\ 
&& w/ & 0.70 & 12.0s\\
&\multirow{2}{*}{w/}& w/o & 0.61 & 1.9s \\
&& w/ & 0.85 & 4.5s \\
  \hline
  \hline
\end{tabular}
\end{adjustbox}
\caption{For smaller open-source LLMs, fine-tuning with synthetic data can enhance the accuracy and latency.}
\label{tbl:2}
\end{table}

\noindent \textbf{Fine-tuning with synthetic data improves accuracy and latency.} \label{fine tuning}
Table \ref{tbl:2} demonstrates the efficacy of our synthetic data generation strategy in enhancing the performance of Mixtral-8x7B and Mistral-7B models through fine-tuning. By integrating synthetic data with CoT methods, we achieve performance levels nearly comparable to those of larger models (85\%, 86\% vs. 89\%, 92\%). This improvement is substantial and justifiable for real-world business applications (e.g., customer support), as smaller models exhibit significantly lower latency. Thus, the fine-tuning with synthetic data not only boosts accuracy but also reduces latency if used with the right-size open-source LLM. The primary reason for the decreased latency with fine-tuning is that the models produce fewer output tokens compared to their non-fine-tuned counterparts due to the fine-tuning data.

However, while CoT enhances the accuracy of the models, it also increases latency across various scenarios, particularly in larger LLMs, which may hinder their use in real-time applications. The smaller Mistral-7B model, do not face this issue, exhibiting latencies nearly tenfold lower than the larger models. This advantage makes fine-tuning smaller models, with CoT, more viable for real-time applications despite the increased latency caused by CoT. 

Based on these results, Model 1 without CoT, Model 1 with CoT and fine-tuned Mistral-7B with CoT are selected for online experiment testing the impact on AMPT. Details of model selection is described in Appendix \ref{app: model selection}.



\begin{table}[ht!]
\centering
\begin{adjustbox}{max width=\textwidth}
\begin{tabular}{c|c}
\hline
\textbf{Suggested Action} & \textbf{AMPT} \\
\hline
\hline
No suggested action & NA (base) \\
\hline
Model 1 w/ CoT & +3\% \\
\hline
Model 1 w/o CoT & -3\% \\
\hline
Fine-tuned Mistral-7B w/ CoT & \textbf{-13\%} \\
  \hline
  \hline
\end{tabular}
\end{adjustbox}
\caption{Compared with other methods, our solution decreases manual processing time by 13\% over baseline.}
\label{tbl:3}
\end{table}

\noindent \textbf{Our solution decreases manual processing time significantly}
Table \ref{tbl:3} illustrates the online evaluation result of manual processing time compared with no suggested action. 
During the online experiment, we found that using a smaller, fine-tuned model (Mistral-7B) with Chain of Thought (CoT) decreased AMPT by 13\%. In contrast, while Model 1 yielded higher quality outcomes, it increased AMPT when used with CoT due to greater latency, and only slightly reduced AMPT (3\%) without CoT. Removing CoT, however, led to a notable decrease in accuracy, resulting in more incorrect actions.

\section{Conclusion}
We propose a novel solution to enhance customer support efficiency by addressing three key challenges: complex internal knowledge, latency in larger LLMs, and scarcity of training data. Our results demonstrate that: (i) the ICA format significantly improves model accuracy, (ii) fine-tuning smaller open-source LLMs can effectively reduce latency and agent work time, and (iii) our synthetic data generation method efficiently created training data, enhancing model performance. This pioneering work not only showcases the application of LLMs in assisting customer support tasks but also sets the stage for future research into reformatting business knowledge across complex domains like legal and finance.

\bibliography{custom}

\appendix

\section{Intermediate Decision Tree for ICA Transformation and Training Data Synthesis}
\label{app: tree}
A workflow can be transformed to a tree structure, with the root node representing the condition on Intent, the internal nodes representing conditions on Contexts, and the leaf nodes representing the Actions. This decision tree can be interconverted with the pseudocode format, where the conditions of root node and internal nodes correspond to if-else clauses and actions of leaf nodes correspond to then-do blocks. This intermediate decision tree will be used in the processes of ICA transformation from original rich text and training data synthesis. 

\section{ICA Transformation from Rich Text} 
\label{app: ica transformation}

Since the HTML containing the original business knowledge in rich text is also in a tree structure, we used HTML parsing tools such as Beautiful Soup \cite{beasutifulsoup} and lxml \cite{lxml} to decompose and extract the text contents from the knowledge while retaining the relationships in the XML tree. Then a binary classifier trained on human labeled data is applied on the extracted content to determine whether it is 1) a condition on intent/context or 2) a description of an action to take. This classification result is used to decide whether the content serves as a leaf node in the intermediate decision tree. The trees are further converted to pseudocode format programmatically and then reviewed, edited and corrected by our human knowledge writers from content and operation team. The entire process takes months for the transformation of the whole knowledge base. However, this process is a one-time effort for the existing legacy knowledge. For new knowledge creation and future updates, knowledge writers can directly create new business logic and edit existing ones in ICA format.






\section{Training Data Synthesis} \label{app: training data}
For the training data synthesis process, we initially created two datasets: one is the pool of the conditions on the intent and context from our internal business knowledge base. The other is a pool of user query and context data from the historical data gathered from loggings of queries and API returned data, with private and sensitive data anonymized or removed. 


With these two datasets, we can synthesize a training instance by the following three steps as illustrated in Figure \ref{fig: 4}:
\begin{itemize}
\item \textbf{Synthesize a matched branch} We randomly sample a user query, a list of context data, and a tree branch consists of one intent condition and multiple context conditions that all can be satisfied by the query and context data.

\item \textbf{Synthesize divergent branches} Upon constructing the matched branch, it remains necessary to generate several divergent branches to facilitate the construction of decision trees. A divergent branch is defined as follows: within a particular branch, there exists more than one node that does not align with the user’s query or associated context data. The generation of a divergent branch may be achieved either through the modification of certain nodes within the matched branch (if the node is the root node, a new tree will be created) or by incorporating an irrelevant branch. 
\item \textbf{Synthesize the CoT} The CoT can help LLM understand the rationale of the action prediction. Given our understanding of which branch is matched and the rationale behind the non-matching status of the other branches during the branch generation phase, we are capable of producing the corresponding reasoning process as the CoT: We construct a list of descriptions of nodes in the matched branch and nodes leading to mismatch of corresponding branch to explain the final action prediction.   

\item \textbf{Convert the synthesized decision trees to ICA format and create the SFT instance} Following the synthesis of all branches in the preceding phase, it is feasible to transform the decision trees into ICA format. We put the synthesized user query, context and ICA in the instruction part while the CoT and action in the label part to create an SFT instance.
 
\end{itemize}

\section{SFT and Model Serving Settings} \label{app: model training}
During training, we use eight A100 GPUs to fine-tune the backbone model with bf16 float precision. Batch size per device is set to 8, training epoch is set to 5, the gradient accumulation step is set to 1 and the max token length is set to 4096. We optimize the model using AdamW optimizer and the learning rate is set to a fixed value of 5e-6. Both LoRA and Full-Parameter fine-tuning are tested and the model with the best performance are selected. 
In the online prediction phrase, for proprietary LLMs, we directly call their interfaces to predict; for open-source models, we use an NVIDIA A100 GPU to serve Mistral-7B, and use eight NVIDIA A100 GPUs to serve other models. The max output token length is set to 512. Additionally, we leverage vllm \cite{kwon2023efficient} to speed up the prediction process. 


\section{Model Selection for Online Experiment} \label{app: model selection}
To minimize dilution of statistical power and ensure the experiment's completion within an acceptable timeframe, we limited the number of experimental groups to four. Models with an offline accuracy (ACC) below 70\% were excluded, based on historical experiments indicating that suggestions with accuracy below this threshold significantly affect the AMPT. Between Mixtral-8x7B and Mistral-7B, we selected fine-tuned Mistral-7B with Chain of Thought (CoT) for its balanced performance—exhibiting high ACC and low AL, and substantially lower serving costs compared to Mixtral-8x7B.
\end{document}